\documentclass[journal,twoside,web]{IEEEtran}
\usepackage{amsmath,amssymb,amsfonts}
\usepackage{algorithm}
\usepackage{textcomp}
\usepackage[table]{xcolor}
\definecolor{lightgray}{gray}{0.9}
\usepackage{booktabs}
\usepackage{lineno}
\usepackage{indentfirst}
\usepackage[noend]{algpseudocode}
\def\algbackskip{\hskip-\ALG@thistlm}
\usepackage{caption}
\usepackage{helvet}
\usepackage{courier}
\usepackage{bm}
\usepackage{epsfig}
\usepackage{enumitem}
\usepackage{rotating}
\usepackage[T1]{fontenc}
\usepackage{booktabs}
\usepackage{subfig}
\usepackage{url,longtable,multirow,morefloats,stfloats, floatflt,cancel,tfrupee}
\makeatletter
\AtBeginDocument{\@ifpackageloaded{textcomp}{}{\usepackage{textcomp}}}
\makeatother
\usepackage{colortbl,xcolor}
\usepackage{pifont}
\usepackage[nointegrals]{wasysym}
\usepackage[format=plain,
            labelfont=it,
            textfont=it]{caption}
\makeatletter
\def\algbackskip{\hskip-\ALG@thistlm}

\@ifundefined{etal}{}{}
\newcommand{\tabincell}[2]{\begin{tabular}{@{}#1@{}}#2\end{tabular}}  
\usepackage{ifxetex}
\ifxetex\else\if@twocolumn\@ifpackageloaded{stfloats}{}{\usepackage{dblfloatfix}}\fi\fi
\usepackage[breaklinks=true,bookmarks=false]{hyperref}

\def\BibTeX{{\rm B\kern-.05em{\sc i\kern-.025em b}\kern-.08em
    T\kern-.1667em\lower.7ex\hbox{E}\kern-.125emX}}

\begin{document}

\title{Domain Adaptive Medical Image Segmentation via Adversarial Learning of Disease-Specific Spatial Patterns} %

\author{Hongwei Li, \IEEEmembership{Student Member, IEEE}, Timo L\"ohr, Anjany Sekuboyina, \\ Jianguo Zhang, \IEEEmembership{Senior member,~IEEE}, Benedikt Wiestler, and Bjoern Menze
\thanks{H. Li and T. L\"ohr make equal contributions to this work}
\thanks{H. Li, T. L\"ohr, A. Sekuboyina, and B. Menze are with Department of Computer Science in Technical University of Munich (e-mails: \{hongwei.li, timo.loehr, anajany.sekuboyina, bjoern.menze\}@tum.de).}
\thanks{B. Wiestler is with University hospital of Technical University of Munich (e-mail: b.wiestler@tum.de).}
\thanks{J. Zhang is with Department of Computer Science and Engineering, Southern University of Science and Technology (e-mail: zhangjg@sustech.edu.cn).}
}

\maketitle
\begin{abstract}
In medical imaging, the heterogeneity of multi-centre data impedes the applicability of deep learning-based methods and results in significant performance degradation when applying models in an unseen data domain, e.g. a new centreor a new scanner. 
In this paper, we propose an unsupervised domain adaptation framework for boosting image segmentation performance across multiple domains without using any manual annotations from the new target domains, but by re-calibrating the networks on few images from the target domain. To achieve this, we enforce architectures to be adaptive to new data by rejecting improbable segmentation patterns and implicitly learning through semantic and boundary information, thus to capture disease-specific spatial patterns in an adversarial optimization. 
The adaptation process needs continuous monitoring, however, as we cannot assume the presence of ground-truth masks for the target domain, we propose two new metrics to monitor the adaptation process, and strategies to train the segmentation algorithm in a stable fashion. 
We build upon well-established 2D and 3D architectures and perform extensive experiments on three cross-centre brain lesion segmentation tasks, involving multi-centre public and in-house datasets. We demonstrate that re-calibrating the deep networks on a few \emph{unlabeled} images from the target domain improves the segmentation accuracy significantly.  
\end{abstract}

\section{Introduction}
\begin{figure}[t]
	\begin{center}
		\includegraphics[width=0.48\textwidth,height=0.33\textwidth]{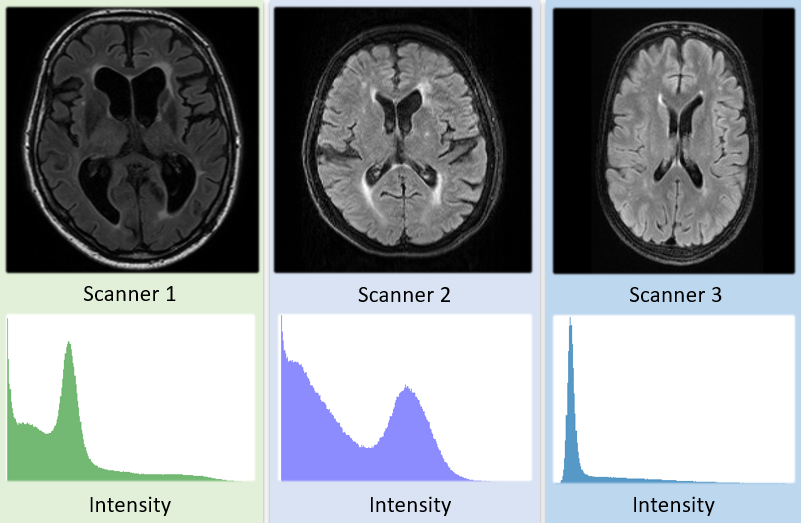}
	\end{center}
    	\caption{Illustration of domain shift in intensity distribution, contrast level and noise level from a MRI FLAIR sequence acquired at different centres. }
	\label{fig:samples}
\end{figure}
Deep learning, in particularly deep convolutional neural networks (CNNs), has achieved remarkable progress in medical image analysis in recent years \cite{litjens2017survey}.  
Medical image segmentation, as an important task in quantifying and structuring image information in biomedical research and clinical practice, plays a crucial role in various applications. Although transformative advancements in segmentation tasks have been achieved by CNNs and their extensions \cite{ronneberger2015u, kamnitsas2017efficient}, most of the supervised learning approaches were built based on a common assumption that training and test data are drawn from the same probability distribution. Thus, those established models are naturally exposed to a \emph{domain shift} in the data encountered at the inference stage \cite{ben2010theory}. As an example, Fig. \ref{fig:samples} depicts the domain shift in terms of intensity distribution, contrast level, and noise level of FLAIR sequences among different scanners.  
Medical images are inherently heterogeneous and complex. Consequently, the performance of computer-aided diagnostic systems may drop rapidly when deployed on arbitrary real-world data domains \cite{glocker2019machine}. 
We refer to data sampled from one distribution to belong to a \emph{domain}. For example, MR images from one scanner with the same imaging protocol belong to one domain whilst the ones acquired in another centre, with a different scanner and a modified imaging protocol belong to another \cite{karani2018lifelong}.  

One way to deal with the heterogeneity of datasets, is sampling from a maximal number of domains and including the data into the training set. Unfortunately, high-quality experts' annotations and clinical follow-up verification for data collected from multiple domains are not always accessible in clinical practice. 
Domain adaptation and transfer learning \cite{pan2009survey}  methods have been studied to generalise established models. Another naive solution is fine-tuning the models learned on the source domain based on annotated data from the target domain. However, 
quality control and approval of the annotation of new cases is manually intensive and would be prohibitive for most existing diagnostic algorithms. 
Therefore, unsupervised domain adaptation (UDA) methods \cite{ganin2014unsupervised} are more feasible in a clinical setting, given that no extra labeled data from the target domain is needed. There are two main streams of work that have addressed UDA: a) Feature-level adaptation, which aims to match the features from source and target domains; b) Image-level adaptation, which transfers the image style from the source domain to the target domain. The first stream requires an empirical selection of the feature level, while the second stream uses a large amount of data to learn the target distribution and synthesize good-quality images. Existing work faces another common drawback: the quality of the adaptation process cannot be validated since labels from the target domain are not available during the evaluation of the validation set.

In this study, we follow two hypothesis to overcome the above drawbacks. First, the disease-specific patterns in segmentation tasks are domain-invariant, i.e., the spatial manifestation of the disease-specific pattern is domain invariant, e.g. the structure or morphology of lesions is invariant to domain shifts. Based on this, we introduce a semantic- and boundary-aware layer to encode the spatial information into the data distribution.
Second, the predicted patterns on the target domain will be iteratively updated during the UDA process and look similar to the source domain, enabling an interpretation of the process even without a validation set. Building on the novel encoding of the spatial information, we further adapt an adversarial learning strategy to enforce the spatial pattern distributions of the source and target domains to be close to each other. Finally, we demonstrate that the variance of the mask differences offers a promising means to monitor the convergence of the UDA process. In summary, the three key contributions of our study are as follows:
\begin{itemize}[noitemsep]
   \item We show that enforcing the cross-domain consistency of spatial patterns-of-interest by following an adversarial UDA learning strategy, improves the generalisation performance of learning-based algorithms, e.g. an segmentation algorithm.
   \item We propose two effective metrics to interpret, monitor and constrain the unsupervised adaptation process without requiring annotated data from the target domain.
   \item We benchmark our UDA method on state-of-the-art architectures and public datasets in cross-centre brain lesion segmentation tasks. We demonstrate significant improvements over baseline in all tasks.
\end{itemize}

\vspace{-0.2cm}
\section{Related Work}
Our work is related to unsupervised domain adaptation, cross-domain image segmentation and continual learning.
\\
\textbf{Unsupervised domain adaptation.} Early studies on UDA focused on aligning the intensity distributions \cite{jager2008nonrigid} or matching the feature space by minimizing the feature distances between the source and target domains \cite{long2015learning,long2016unsupervised}.  Recently, with the advances of generative adversarial network (GAN) and its extensions \cite{goodfellow2014generative,arjovsky2017wasserstein}, the latent features across different domains can be implicitly aligned by adversarial learning. Y. Ganin \emph{et al.} \cite{ganin2016domain} proposed to learn domain-invariant features by sharing weights between two CNNs.
 E. Tzeng \emph{et al.} \cite{tzeng2017adversarial} introduced a unified framework in which each domain is equipped with a dedicated encoder before the last softmax layer. 
However, as commented in \cite{zhang2017curriculum}, the above UDA methods for classification tasks cannot work well to address dense segmentation tasks \cite{hoffman2016fcns}, because the mapping functions from image space to label space may differ in source and target domains due to the domain shift. In medical imaging, several works \cite{ouyang2019data, orbes2019knowledge} have been proposed to tackle UDA in different scenarios. Yet, there exists no metric available to interpret and constrain UDA process with only data from the source domain. 
\\
\textbf{Cross-domain image segmentation.}
In contrast to many prior studies from computer vision, medical image segmentation is a highly-structured prediction problem. 
The existing work can be divided into two main streams: \emph{feature-level} adaptation and \emph{image-level} adaptation.
In the feature level, K. Kamnitsas \emph{et al.} \cite{kamnitsas2017unsupervised} made an early attempt to perform UDA for brain lesion segmentation which aimed to learn domain-invariant features with a domain discriminator. The cross-modality segmentation problem with a large domain shift is addressed in \cite{dou2018unsupervised} in which specific feature layers are fine-tuned and an adversarial loss is used for supervised feature learning. In comparison to our proposed approach, it should be noted that the loss functions aim to match only image features but not the resulting segmentation maps. Moreover, it only adjusts the feature transform, but not the segmentation outcome itself. 
Existing studies \cite{zhao2018supervised,huo2018adversarial,bousmalis2017unsupervised} demonstrate that the image-level adaptation brings improvements in pixel-wise predictions on a target domain. In this direction, the domain shift problem is addressed at the input level by providing target-like synthetic images. A recent work \cite{chen2019synergistic} combines feature-level and image-level adaptation and achieves state-of-the-art results on cross-modality segmentation. In \cite{zhang2020generalizing}, a data augmentation technique stacking several classical processing methods was proposed to generalise networks to unseen domains. \\
\textbf{Continual learning.} Whilst adapting the model to a target domain, we expect that the performance on the source domain remains the same without catastrophic forgetting. I. Goodfellow \emph{et al.} \cite{goodfellow2013empirical} studied \emph{dropout} techniques to regularize the training and avoid catastrophic forgetting. F. Ozedemir \emph{et al.} proposed to select representative data for image segmentation \cite{ozdemir2018learn}.  As an early attempt to learn from multi-domain MRI data, N.  Karani \emph{et al.} \cite{karani2018lifelong} proposed to learn a normalisation layer to adapt to different domains. 
We propose to expand the capacity of well-trained networks and maintain the performance on the original source domains.

The innovation of this work is fundamentally different from existing methods for two reasons. First, we do not assume that the features of the source and target domains should be aligned. 
As mentioned in \cite{zhang2017curriculum}, the assumption that feature alignment in source and target domains without carefully considering the structured labels, becomes less likely to hold. 
Second, our work can handle few-shot scenario in the target domain. For the \emph{image-level} adaptation ones, the quality of the synthetic target-like images is not guaranteed without large amount of data from the target domain, especially when the region of interest is tiny (e.g. small brain lesions) in our study. 

\section{Method}
\begin{figure*}[t]
	\begin{center}
		\includegraphics[width=0.91\textwidth,height=0.44\textwidth]{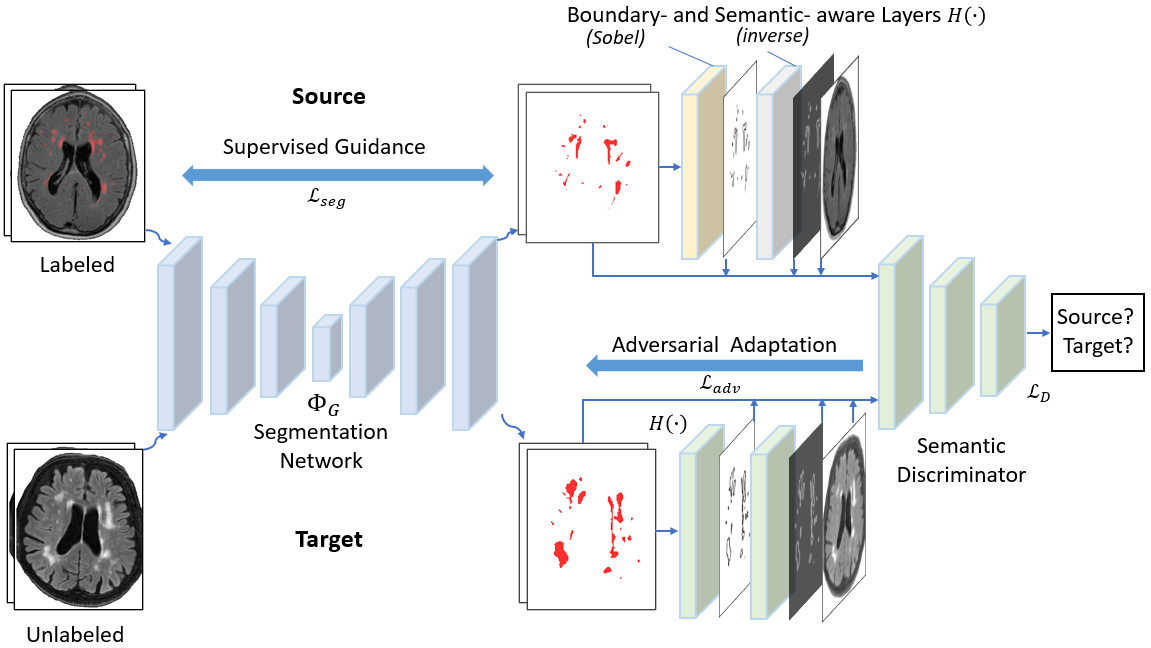}
	\end{center}
    	\caption{Overview of the proposed efficient unsupervised domain adaptation framework, consisting of a segmentation model $\Phi_{G}$ and a semantic discriminator $D$. The semantic distributions of source and target domains are driven to be similar through adversarial learning. The weights of the two segmentation models are shared and they are trained on both source and target domains in a supervised and an adversarial manner respectively. The semantic discriminator takes the raw images, semantic masks, edge maps and inverse maps to implicitly learn the \emph{domain-invariant} disease-specific spatial pattern.}
	\label{fig:framework}
\end{figure*}

An overview of the proposed method is shown in Fig. \ref{fig:framework}. Given a segmentation model pre-trained on the source domain, our objective is to adapt this model to an unseen target domain. We developed a framework using a semantic discriminator which enforces a similar image-to-label mapping in the source and target domain. This encourages the generator (i.e. segmentation network) to produce probable segmentation.

\subsection{Problem Definition, Assumption and Notation}
Let $\mathcal{X}$ denote an input image space and $\mathcal{Y}$ a segmentation label space. We define a domain to be a joint distribution $\mathbb{P}_{XY}$ on $\mathcal{X}$ $\times$ $\mathcal{Y}$. 
Let $\mathfrak{P}_{S}$ and $\mathfrak{P}_{T}$ denote the set of source domains and target domains.
We observe $N$ source domains $\mathcal{S} = \{S^{i}\}_{1}^{N}$, where $S^{i} = \{(x_{k}^{(i)}, y_{k}^{(i)})\}_{k=1}^{n_{i}}$ is sampled from $\mathfrak{P}_{S}$ containing ${n_{i}}$ samples,
and $M$ target domains $\mathcal{T} = \{T^{j}\}_{1}^{M}$, where $T^{j} = \{x_{k}^{(j)}\}_{k=1}^{m_{j}}$ is sampled from $\mathfrak{P}_{T}$.
Notably, the samples from the target domain do not contain any ground-truth labels.

We consider two mapping functions from image spaces to label spaces. $\Phi_{S}$ : $\mathcal{X}_{S} \rightarrow \mathcal{Y}_{S}$ is the mapping learnt in the source domain, and $\Phi_{T}$ : $\mathcal{X}_{T} \rightarrow \mathcal{Y}_{T}$ is the one learnt in the target domain (if labels are available).
We assume that ideally the two mappings from image space to label space are \emph{domain-invariant} in image segmentation tasks, i.e., $\Phi_{S} = \Phi_{T}$. In other words, the spatial pattern distribution of the segmentation is stable across different domains, e.g., the presence of brain lesions in the MR images.
The goal of UDA is to learn a generalised domain-invariant segmentation model $\Phi_{G}$ given \{$\mathcal{S}$, $\mathcal{T}$\} such that $\Phi_{G}$ approximates $\Phi_{S}$ and $\Phi_{T}$. We aim to learn $\Phi_{G}$ using $\Phi_{S}$ as a reference since $\Phi_{T}$ is not observed. 
The network in the source domain $\Phi_{S}$ can be learnt in a fully supervised fashion given $\mathcal{S}$. Before domain adaptation, $\Phi_{S}$ does not generalise to $\mathcal{T}$ and can be an initialisation of $\Phi_{G}$. Since the domain-invariant model $\Phi_{G}$ is expected to generalise on both $\mathcal{S}$ and $\mathcal{T}$ in our setting, the training of $\Phi_{G}$ is partly supervised by $\mathcal{S}$ to avoid catastrophic forgetting on the source domain during the UDA process.

\subsection{Semantic- and boundary-aware layer}
Based on the hypothesis that disease-specific spatial patterns are domain-invariant, we claim that the combination of spatial information like semantic and boundary information is crucial for domain adaptation. Especially the boundary-aware input showed its effectiveness in previous studies \cite{ding2019boundary,shen2017boundary}. Therefore, we introduce a semantic- and boundary-aware layer which incorporates semantic and boundary information into the distribution by spatially concatenating the image, semantic masks and edge maps as a part of the discriminator input. We introduce the \emph{Sobel layer} as shown in Fig. \ref{fig:framework} to detect the boundary in predicted masks using \emph{Sobel} operators \cite{gao2010improved}. Since the labels of the background and pathology are often highly unbalanced, especially in brain lesion segmentation task, we further develop an \emph{inverse layer} which inverts the maps in order to reinforce the learning on pathology regions. This enhancement facilitates the overall model learning.

Let $P$ denote the probability maps, meaning the prediction and ground-truth maps, $f_1$ and $f_2$ denote the Sobel operations for two directions, and $J$ denote the all-ones matrix with the same size with $P$, the boundary maps are defined as $f_1(P)$ and $f_2(P)$. The inverse label maps are defined as: $J-P$.
Thus the semantic- and boundary-aware layers $H(P)$ concatenate multiple maps as its input:
    \begin{equation}
    \emph{$H(P)$} = [\emph{$P$, $J-P$,$f_1(P)$, $f_2(P)$,  $J-f_1(P)$, $J-f_2(P)$}]
    \end{equation}
The effectiveness of the proposed semantic and boundary-aware layers will be presented in Section \ref{ablation}.

\subsection{Adversarial Domain Adaptation}
\subsubsection{Backbone ConvNets for segmentation} One basic component of the proposed approach is a fully convolutional neural network (ConvNets) for image segmentation. With the $N$ labeled samples from the source domain, supervised learning was conducted to establish a mapping $\Phi_{S}$ from the input image space $\mathcal{X}_{S}$ to the label space $\mathcal{Y}_{S}$. We borrowed the top-performing U-shape 2D and 3D architectures from \cite{li2018fully, wang2017automatic} and adopted the same configuration for all meta-parameters. Other architectures can certainly be used in our framework. The parameters of the network are learnt by iteratively minimizing a segmentation loss using stochastic gradient descent. The segmentation loss function is a linear combination of Dice loss \cite{milletari2016v} and cross-entropy loss, formulated as
\begin{equation} \label{loss_function}
    \begin{aligned}
    \mathcal{L}_{seg} = - \lambda\frac{2\sum_{i=1}^n  y_{i}p_{i} + s}{\sum_{i=1}^n y_{i}^2  + \sum_{i=1}^n p_{i}^2  + s}  \\
    - (1-\lambda) \sum_{i = 1}^{n}(y_{i}\log(p_{i})+(1-y_{i})\log(1-p_{i}))\\
    \end{aligned}
\end{equation}

where $s$ is the smoothing factor to avoid numerical issues, $y_{i}$ and $p_{i}$ are the ground-truth label and the probability of the prediction for the $i_{th}$ voxel respectively. Notably the loss function can be a multi-class version depending on the segmentation task.
We use the same ConvNet for both source and target domains. We claim that a deep ConvNet offers large modeling capacity and can learn effective representations on diverse data from multiple domains as observed from recent segmentation benchmarks \cite{bakas2018identifying,kuijf2019standardized,zhuang2019evaluation}.
Given the modelling capacity, the goal of this work is to learn a domain-invariant segmentation model across multiple similar domains. 
\\
\subsubsection{Discriminator and adversarial learning} 

The goal of learning a domain-invariant $\Phi_{G}$ is equal to minimizing the distance between $\Phi_{G}$ and $\Phi_{S}$.
Inspired by \cite{ganin2016domain}, we adopt an adversarial network including a generator $G$ which performs segmentation for given input images, and a discriminator $D$ that evaluates whether the image-to-label mapping is the same as $\Phi_{S}$ or not, thus pushes $\Phi_{G}$ to be close to $\Phi_{S}$.
In other words, the mapping $\Phi_{S}$ from source image space to ground-truth label space is treated as '\emph{expert knowledge}' while the mapping $\Phi_{G}$ is treated as '\emph{machine knowledge}'. However, modelling of $\Phi_{G}$ faces the challenge of measuring the similarity between mappings which is not straightforward to compute. Meanwhile, $\Phi_{S}$ is difficult to be directly formulated.
We parameterize these functions by using a deep neural network $D$ to map the samples from \{$\mathcal{S}$, $\mathcal{T}$\} along with spatial pattern information to a latent feature space and discriminate the underlying distribution to be same or not. The optimization of $D$ and $\Phi_{G}$ can be formulated as:

\begin{equation}
    \begin{aligned}
    \min_{\Phi_G}\max_D\mathcal{L}_{adv}(\mathcal{S}, \mathcal{T}, D, \Phi_{G}) = \mathbb{E}_{(x, y)\sim\mathcal{S}}[\log D(x,H(y))]+\\
    \mathbb{E}_{x\sim\mathcal{T}}[\log (1-D(x, H(\Phi_{G}(x))))]\\
    \end{aligned}
\end{equation}

However, this objective can be problematic since during the early training stage the discriminator converges quickly, causing issues of vanishing gradients. It is typical to train the generator with the standard loss function with partly inverted labels \cite{goodfellow2014generative}. In particular, some ground-truth maps are labeled as 'fake' and some predicted maps are labeled as ground-truth. This regularizes the segmentation network to avoid over-fitting. Consequently, the optimization is split into two independent objectives, one for the discriminator and one for the generator:

\begin{equation} \label{discriminator}
    \begin{aligned}
    \min_{D}\mathcal{L}_{adv_{D}}(\mathcal{S}, \mathcal{T}, D, \Phi_{G}) =
    - \mathbb{E}_{(x, y)\sim\mathcal{S}}[\log D(x,H(y))]-\\
    \mathbb{E}_{x\sim\mathcal{T}}[\log (1-D(x, H(\Phi_{G}(x))))]
    \end{aligned}
\end{equation}

\begin{equation} \label{generator}
    \begin{aligned}
    \min_{\Phi_{G}}\mathcal{L}_{adv_{M}}(\mathcal{S}, \mathcal{T}, D, \Phi_{G}) =  - \mathbb{E}_{x\sim\mathcal{T}}[\log (D(H(\Phi_{G}(x))))]\\
    \end{aligned}
\end{equation}

Notably, this two-stage optimization provides stronger gradients to update $\Phi_{G}$ and uses inverted labels to update the generator in Eq.\ref{generator}. In addition, catastrophic forgetting of source domain information is avoided by presenting samples from both $S$ and $T$ to the segmentation model during the domain adaptation process. \\

\subsection{Monitoring Metrics}
Since we assume that no ground-truth labels from the target domain are available, the convergence of the UDA process cannot be directly observed. We introduce two metrics to interpret and constrain the UDA process. The idea is that the segmentation model $\Phi_{G}$ generates a starting-point prediction on the target domain after pre-training on the source domain. When performing UDA, the prediction updates towards the ground-truth which causes the increase of the difference between the updated prediction and the initial segmentation. We make an assumption that the difference between the current and the initial segmentation mask is expected to show stable at some iterations.

Given a sample $x$ from target domain, let $A_0$ denote the initial mask predicted by the pre-trained model before UDA, formulated as:
$A_0 = \Phi_{G}^{0}(x)$.
Let $A_i$ be the mask predicted by $\Phi_{G}^{i}$ at $i_{th}$ iteration during the UDA process, formulated as:
$A_i = \Phi_{G}^{i}(x)$.
Euclidean distance of two masks was used as one of the monitoring metrics to measure the closeness, formulated as:
\begin{align}
    d_i = \vert\vert A_i - A_0\vert\vert
\end{align}
where $\vert\vert \cdot \vert\vert$ represents the Euclidean distance.

To measure the stability, we further compute the variance $\sigma_d^2$ of the mask differences in an interval of $k$ iterations, which can be formulated as:
\begin{align} \label{variance}
    \sigma_d^2 = \frac{\sum_{j=m+1}^{m+k} (d_j - \mu)^2}{k}
\end{align}

where $\mu$ is the average of the mask differences in an interval of $k$ iterations. When $\sigma_d^2$ is below a certain threshold $\epsilon$, it indicates that the prediction behaves stable. Two metrics can be computed as the stopping criterion for the learning process.

\subsection{Training Strategies}
In this section, we propose a training strategy that was observed to converge stably. We define an end-to-end two-stage training algorithm (cf. Algorithm \ref{euclid}). First, we train a segmentation model in a fully supervised fashion on the source domains. Second, we enforce the same model to be domain-invariant in an adversarial fashion while maintaining the performance on the source domains.
\begin{algorithm}
    \caption{Unsupervised domain adaptation process}\label{euclid}
    \hspace*{\algorithmicindent} \textbf{Input:}  $\mathcal{S}$ from source domain, $\mathcal{T}$ from target domain, number of epochs $m$, stopping threshold $\epsilon_{1}$, $\epsilon_{2}$\\
    \hspace*{\algorithmicindent} \textbf{Output:} segmentation model $\Phi_G$,  discriminator $\mathcal{D}$ \\
    \hspace*{\algorithmicindent} \textbf{Initialise} $\Phi_G$ and $\mathcal{D}$
    \begin{algorithmic}[1]
    \Procedure{Pre-training}{}
    \State get batches from $\mathcal{S}$ and update~$\Phi_G$ by Eq. (\ref{loss_function})
    \State compute $A_{0} = \Phi_G(\mathcal{T})$
    \EndProcedure

    \Procedure{Domain adaptation}{}
    \State get batches with domain labels from $\mathcal{S}$ and $\mathcal{T}$
    \While {$\sigma_d^2 < \epsilon_{1}$ and $d_{j} > \epsilon_{2}$}
    \State $\{d_j\}=[~], \sigma_d^2 = 0$, j = 0
    \While{j < n }

    \State update~$\Phi_G$ and $D$ with a batch from $\mathcal{S}$ 
    \State update~$\Phi_G$ and $D$ with a batch from $\mathcal{T}$  
    \State only update~$\Phi_G$ with a batch from $\mathcal{T}$  
    \State compute $d_j = \vert\vert \Phi_G(\mathcal{T}) - A_{0}\vert\vert$
    \State j = j+1
    \EndWhile
    \State compute $\sigma_d^2$ by Eq. \ref{variance}
    \EndWhile
    \Return $\Phi_G$
    \EndProcedure

    \end{algorithmic}
    \end{algorithm} \label{algorithm_1}

\section{Experiments}

\begin{table*}[htpb]
  \caption[table: Data Properties]{Data Characteristics of the public datasets and internal dataset of three segmentation tasks. TR/TE/TI are imaging parameters from specific imaging protocols. }\label{tab:data_prop}
  \centering
   \begin{tabular}{c| c c l l c c c}
    \hline
      Tasks&Centres~&~Scanner&Voxel Size (mm$^{3}$) &Volume Size & Modality & TR/TE/TI (ms)& Num. \\
    \hline
      White Matter&Utrecht~&~3T Philips Achieva&$0.96\times0.95\times3.00$ & $240\times240\times48$ &FLAIR, T1& 11000/125/2800 & 20 \\
      Hyperintensities&Singapore~&~3T Siemens TrioTim&$1.00\times1.00\times3.00$ & $252\times232\times48$ &FLAIR, T1& 9000/82/2500 &20 \\
      &Amsterdam~&~3T GE Signa HDxt&$0.98\times0.98\times1.20$ & $132\times256\times83$ &FLAIR, T1& 8000/126/2340& 20 \\
      \hline
      MS lesion&Munich~&~3T Philips Achieva&$1.00\times1.00\times1.00$ & $170\times240\times240$ &FLAIR, T1& -& 50 \\
      &JHU~&~3T Philips Intera&$1.00\times1.00\times1.00$ & $181\times181\times217$ &FLAIR, T1 &10.3/6/835& 20 \\
      \hline
      Brain Tumor&Upenn~& - &$1.00\times1.00\times1.00$ & $170\times240\times240$ &FLAIR, T1, T2, T1-c& -& 92 \\
      &Others&-&$1.00\times1.00\times1.00$ & $170\times240\times240$  &FLAIR, T1, T2, T1-c&-& 50 \\
      \hline
      
     \hline
  \end{tabular}
\end{table*}
\subsection{Datasets and Evaluation Metrics} 
\subsubsection{Datasets} We validate our method on three image segmentation tasks, acquired across more than seven centres. \\
\textit{\textbf{Task 1:}} White matter hyperintensities (WMH) segmentation. We use the public datasets of MICCAI White Matter Hyperintensities Segmentation Challenge 2017 \cite{kuijf2019standardized} including three centres. Each centre contains 20 subjects and FLAIR\&T1 MRI modalities; \\
\textit{\textbf{Task 2:}} Multiple sclerosis (MS) lesion segmentation. We use an in-house MS dataset consisting of 50 subjects treated to be the source domain and perform UDA on a recent public dataset (ISBI-2015) \cite{carass2017longitudinal} including 20 subjects and multiple modalities treated to be the target domain. Only FLAIR and T1 are used since they are in common.\\
\textit{\textbf{Task 3:}} Brain tumor segmentation. We use the public MICCAI-BraTS2019 \cite{bakas2018identifying,menze2014multimodal} glioma dataset for evaluation. In this, we have the information of which samples are from UPenn and which are not. We choose UPenn samples (92 subjects) to be from the source domain and randomly selected 50 subjects from the remaining samples to be from an unknown target domain. Table \ref{tab:data_prop} presents the characteristics of the three datasets used.

\subsubsection{Evaluation metrics}
For evaluating the performance, we use five evaluation metrics.  
Given the ground-truth segmentation mask $G$ and the generated mask $P$ by the segmentation model $\Phi_{G}$, the evaluation metrics are defined as follows. \\
\textit{\textbf{Dice similarity coefficient (DSC):}} $DSC = \frac{2(G\cap P)}{\vert G \vert + \vert P \vert}$ \\
\textit{\textbf{Hausdorff Distance (H95):}} \\
$H95 = max\{\sup_{x \in G}\inf_{y \in P} d(x,y),~ \sup_{y \in G}\inf_{x \in P}d(x,y)\}$,
\\
where $d(x,y)$ denotes the distance between $x$ and $y$, \emph{sup} denotes the supremum and \emph{inf} the infimum. 
For robustness, a $95^{th}$ percentile version instead of the maximum distance ($100^{th}$ percentile), was used.\\
\textit{\textbf{Absolute Volume Difference (AVD):}}
$V_G$ and $V_P$ denote the volume of lesion regions in $G$ and $P$ respectively. The AVD is defined in percentage as:
$AVD = \frac{\vert V_G - V_P \vert}{V_G}$. \\
\textit{\textbf{Lesion-wise Recall:}}
$N_G$ denotes the number of individual lesion regions in $G$, while $N_c$ denotes the number of correctly detected lesions in $P$.
$Recall = \frac{N_c}{N_G}$. \\
\textit{\textbf{Lesion-wise F1-score:}}
$N_c$ denotes the number of correctly detected lesions in $P$, while $N_f$ denotes the wrongly detected lesions in $M$. The F1-score is defined as: $F1 = \frac{N_c}{N_c + N_f}$.

For the first task (WMH), we use all the five metrics according to the challenge setting; For the second task (MS), only Dice score, H95 and AVD are used as they are the main evaluation metrics in existing literature. For the third one (brain tumor), we evaluate Dice of three tumor tissues according to the challenge setting. 
\vspace{-0.2cm}
\subsection{Implementations}
\subsubsection{Image preprocessing}
For tasks 1 and 2, we use 2D axial slices of both FLAIR and T1 sequences for training. All images are cropped or padded to a uniform size of 200 by 200 pixels. Then the voxel intensity is normalised with \emph{z-score} normalisation subject-wise. We use data augmentation (rotation, shearing and scaling) to achieve the desired invariance in the training stage. For task~3, we perform bias field correction to improve image quality. We randomly extract cubes from the 3D volumes. Each cube is with a size of 76$\times$96$\times$96 and is normalised with \emph{z-score} normalisation. 

\begin{table*}[htpb]
  \caption{Results on three cross-centre WMH segmentation tasks and comparison with state-of-the-art methods. The values are calculated by averaging the results on the target dataset. {Baseline} denotes the performance without using adaptation.}\label{tab:results_1}
  \centering
   \begin{tabular}{c c c c c c |c |c}
    \hline
      Conditions  ~&~Dice score&H95$\downarrow$ (mm)  & AVD$\downarrow$ & Lesion Recall& Lesion F1 & {\tabincell{c}{p-value$_{Dice}$\\ $[$ours \emph{vs.} others$]$}} & {\tabincell{c}{p-value$_{H95}$\\ $[$ours \emph{vs.} others$]$}} \\
    \hline
    \emph{\textbf{U. + A. $\rightarrow$ S.}} &&&&&&\\
      Baseline &~0.682 & 9.22& 45.95& 0.641 & 0.592 & $<$0.001 & $<$0.001 \\
      U-Net Ensembles \cite{li2018fully}&~{0.703} & {8.83}& {37.21}& {0.672} & {0.642}& $<$0.001 & 0.008\\
      CyCADA \cite{hoffman2017cycada}&~{0.452} & {15.23}& {67.13}& {0.462} & {0.344} & $<$0.001 & $<$0.001\\
      BigAug \cite{zhang2020generalizing} &~{0.711} & {8.25}& {35.41}& {0.691} & {0.651}& $<$0.001 & 0.012 \\
      Ours (with a few shots) &~{0.780} & {7.54}& {24.75}& {0.666} & \textbf{0.657} & 0.325 & 0.599\\
      Ours (with full set) &~\textbf{0.782} & \textbf{7.51}& \textbf{22.14}& \textbf{0.754} & {0.649} & - & - \\
     \hline
      \emph{\textbf{U. + S. $\rightarrow$ A.}} &&&&&&\\
       Baseline &~0.674 & 11.51& 37.60& 0.692 & 0.673& 0.002 & $<$0.001\\
       U-Net Ensembles \cite{li2018fully}&~{0.694} & {9.90}& {31.01}& {0.720} & {0.691} & 0.008 & 0.002\\
       CyCADA \cite{hoffman2017cycada}&~{0.412} & {18.21}& {89.23}& {0.402} & {0.292}& $<$0.001 & $<$0.001\\
       BigAug \cite{zhang2020generalizing} &~{0.691} & {9.77}& {30.64}& {0.709} & {0.704} & 0.012 & 0.008\\
      Ours (with a few shots) &~{0.733} & 7.90& \textbf{16.01}&{0.785} & {0.725} & 0.530 & 0.357\\
      Ours (with full set)&~\textbf{0.737} & \textbf{7.53}& {30.97}&  \textbf{0.841} & \textbf{0.739}& - & - \\
     \hline
      \emph{\textbf{A. + S. $\rightarrow$ U.}} &&&&&&\\
       Baseline &~0.430 & 11.46& 54.84& 0.634 & 0.561 & $<$0.001 & $<$0.001 \\
       U-Net Ensembles \cite{li2018fully}&~{0.452} & {10.38}& {50.33}& {0.652} & {0.565} & $<$0.001 & $<$0.001\\
       CyCADA \cite{hoffman2017cycada}&~{0.422} & {13.91}& {77.45}& {0.544} & {0.385}& $<$0.001 & $<$0.001\\
      BigAug \cite{zhang2020generalizing} &~\textbf{0.534} & \textbf{9.49}& \textbf{47.46}& {0.643} & \textbf{0.577} & 0.262 & 0.470\\
      Ours (with a few shots) &~{0.489} & 11.02& 57.01&{0.639} & {0.533}&0.008 & 0.002\\
      Ours (with full set) &~{0.529} & {10.01}& {54.95}&  \textbf{0.652} & {0.546} & - & - \\

    \hline
\vspace{-0.1cm}
  \end{tabular}
\end{table*}
\vspace{-0.1cm}
\subsubsection{Network architectures and parameters setting}
The generator can be any state-of-the-art segmentation network. In this work, we employ the architectures proposed in \cite{li2018fully, wang2017automatic}. For task 1 and 2, we use a 2D architecture. The generator takes the concatenation of FLAIR and T1 image as a two-channel input and follows a U-net structure. A combination of convolutional and max-pooling layer downsamples the input data before the segmentation mask is produced by several upsampling layers. Additionally, skip connections between layers at the same level create a stronger relation between input and output. The output is a single-channel probability map with pixel-wise predictions for the input image. For task~3, the input is the concatenation of FLAIR, T1, T2 and T1-c volumes. The output is a multi-channel probability map with voxel-wise predictions. 

The discriminator module is a convolutional network (either 2.5-D or 3.5-D including the semantic and boundary channels)  which aims to discriminate between the spatial distributions of the source and target domains. For task 1 and 2, the discriminator takes a seven-channel input consisting of the paired FLAIR \& T1 images, the segmentation mask, the mask's inverse-label maps, two \emph{Sobel} edge maps and the edge maps' inverse-label maps. The inverse map and edge map introduce semantic and boundary information of the critical structure which helps the network to evaluate the quality of the segmentation. This is necessary because in tasks 1 and 2 the interested areas are very small. We use a PatchGAN \cite{isola2017image} architecture with a small patch size of 10 $\times$ 10 as the output. The discriminator model is trained with domain labels using a cross-entropy loss. 
For task~3, the architecture described above is retained but with 3D convolutions instead of 2D. The input for the 3D discriminator is 5$\times$8$\times$8. 

An Adam optimiser \cite{kingma2014adam} is used for stochastic optimization. The learning rates for the segmentation network, discriminator model and adversarial model are set to 0.0002, 0.001 and 0.0002 respectively chosen empirically observing the training stability. The model is trained on a NVIDIA Titan V GPU with 12GB memory. The batch size is set to 30 for the task~1\&2 and 5 for the task~3 considering GPU memory. 
\vspace{-0.2cm}
\subsection{Results}
\subsubsection{Training and Baseline Methods} ~\\
A baseline model (i.e. segmentation part) is trained and optimised on the source datasets, referred to as $\mathcal{S}$ to establish a lower bound performance. It is then tested on the subjects from a target domain $\mathcal{T}$. We use a validation set (20\% of the source training set) to optimise the hyper-parameters and to guarantee that we compare with a strong baseline. 
To demonstrate the efficiency of our method, we run the algorithm considering two conditions of the target domain:
(a) Using only \textbf{a few shots} of target images: In this scenario, we use the data from $\mathcal{S}$ and only a tiny subset (i.e. one scan, \textasciitilde 1-2\% of the target images) of the target domain images without any annotation from the target domain.
(b) Using \textbf{the full set} of the target domain: Here, we use the data from $\mathcal{S}$ and the full set of the target domain images without any annotation from the target domain.


\begin{figure*}[t]
	\begin{center}
		\includegraphics[width=1.0\textwidth,height=0.54\textwidth]{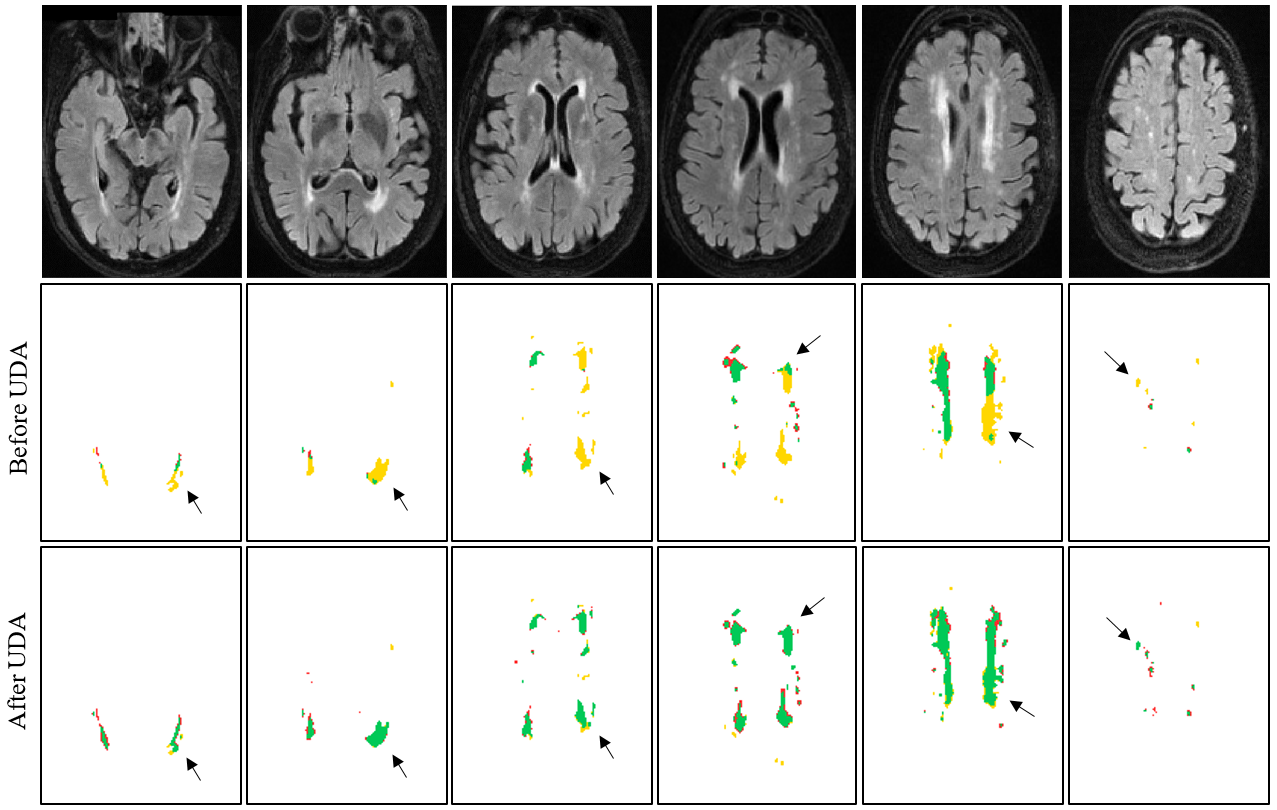}
	\end{center}
	\vspace{-0.1cm}
    	\caption{From left to right: results on six axial slices of the same subject. From top to bottom: FLAIR axial-view images, the segmentation results before UDA, the segmentation results using the proposed method. Green color indicates overlap between the segmentation result and the ground truth masks; red color false positives; gold color false negatives. (Best viewed in color)}
	\label{fig:slices}
\vspace{-0.1cm}
\end{figure*}

\subsubsection{Domain Adaptation on WMH Segmentation} 
\begin{figure}[t]
	\begin{center}
		\includegraphics[width=0.38\textwidth,height=0.21\textwidth]{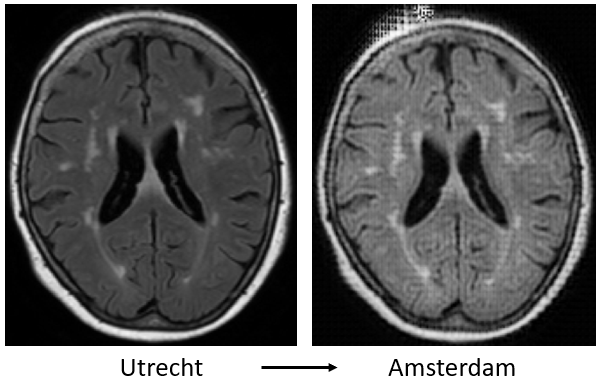}
	\end{center}
    	\caption{The translation from \emph{Utrecht} to \emph{Amsterdam} using CycleGAN \cite{zhu2017unpaired}. We observed that it introduce noise in the synthetic domain due to the lack of training data. Image-translation based method is likely to fail when the region of interests (i.e. lesions) is small. 
    	}
    \label{fig:failure}
\end{figure}

~\\
\textit{\textbf{Utrecht + Amsterdam $\rightarrow$ Singapore:}}
For this setting, we take centre 1 and 3 (\emph{Utrecht} and \emph{Amsterdam}) as the source domain which leaves centre 2 (\emph{Singapore}) for the target domain.
Table \ref{tab:results_1} presents the final results of our approach on the target dataset and compares with baseline and state-of-the-art U-Net ensemble method \cite{li2018fully}. Notably \cite{li2018fully} is the top-performing algorithm on cross-scanner segmentation as analyzed in \cite{kuijf2019standardized}. CyCADA \cite{hoffman2017cycada} is an image-level CycleGAN-based adaptation method. We find that CyCADA achieves poor results because of the low image quality in the synthesized target domain. Fig. \ref{fig:failure} shows a failure example of CyCADA
's synthesized result. Our method significantly improves the segmentation performance on the target dataset after domain adaptation (ours \emph{vs.} baseline, p-value < 0.001). We observe that our method with a few shots of target images achieves a similar Dice score compared to using the full set of the target domain (78.0\% \emph{vs.} 78.2\% \cite{li2018fully}). When using the full set, our method achieves a promising performance close to the fully-supervised result \cite{li2018fully} for the Dice score (78.2\% \emph{vs.} 80.3\%) and lesion-wise recall (75.4\% \emph{vs.} 76.1\%).\\
\textit{\textbf{Utrecht + Singapore $\rightarrow$ Amsterdam:}}
In this experiment, we take centre 1 and 2 (\emph{Utrecht} and \emph{Singapore}) as the source dataset whilst centre 3 (\emph{Amsterdam}) represents the target dataset. Similarly, we observe that using a few shots of target images can significantly improve the segmentation results on target domain. When using the full set, our method further boost the performance, e.g. in lesion-wise recall, from 78.5\% to 84.1\%. The performance in AVD in Table \ref{tab:results_1} decreases after using the full set whilst the Dice stays stable. This indicates that the algorithm encourages the network to produce reasonable prediction based on spatial patterns of the lesions. \\
\textit{\textbf{Amsterdam + Singapore $\rightarrow$ Utrecht:}} Lastly, we take centre 2 and 3 (\emph{Amsterdam} and \emph{Singapore}) as the source dataset whilst centre 1 (\emph{Utrecht}) represents the target dataset. We observe that the baseline performance is relatively poor (43.0\% Dice). This demonstrates that the domain shift between the source (centre 2 and 3) and target (centre 1) domain is in this case greater than in the first two experiments and the performance drastically drops when the target domain is unseen. After UDA, we improve the Dice by nearly 10\% (43.0\% \emph{vs.} 52.9\%). Still, there is room for improvement in this scenario. The state-of-the-art method \emph{bigAug} \cite{zhang2020generalizing} employing advanced data augmentation techniques and improve the generaliability to unseen domains. However, we find that the performance differences between ours and \emph{bigAug} are not significant (Wilcoxon rank-sum tests) as shown in Tab. \ref{tab:results_1}. 

\begin{table}[ht]
  \caption[table: Results MS]{Results on a cross-centre segmentation task for Multiple Sclerosis (MS) data. The values are calculated by averaging the results on the target dataset. We compare our method with the performance before using domain adaptation (denoted by \emph{baseline}). Statistical analysis (ours with full set \emph{vs.} others) is performed on Dice scores.}\label{tab:results_2}
  \centering
   \begin{tabular}{c c c c c}
    \hline
      Method  ~&~Dice score&H95$\downarrow$ & AVD$\downarrow$ & p-value$_{Dice}$\\
    \hline
      Baseline &~0.477 & 19.13& 78.86 & <0.001\\
      Ours (with a few shots) &~{0.492} & {18.06}& {73.83} & <0.001\\
      Ours (with full set) &~\textbf{0.543} & \textbf{13.93}& \textbf{49.92} & -\\
    \hline
  \end{tabular}
\end{table}

\subsubsection{Domain Adaptation on MS Lesion Segmentation}~\\
\textit{\textbf{Munich $\rightarrow$ JHU:}}
In this experiment, we test our approach on two datasets from the multiple sclerosis domain. The Munich dataset, collected in-house, serves as the source dataset while the public JHU dataset is taken as our target domain. Table \ref{tab:results_2} presents the results of our unsupervised domain adaptation in comparison to the baseline performance. Compared to the baseline, our method improves the results even when using only a few shots of target images. Furthermore, when using the whole target dataset the results improve significantly (50.1\% \emph{vs.} 55.2\%). The performance underlines the success of our approach and shows the validity across different domains.
\textit{\textbf{JHU $\rightarrow$ Munich:}}
Due to the few number of scans in the source domain, the segmentation network overfits heavily on the source dataset during the pre-training stage. This prevents a successful adversarial learning during the domain adaptation.

\subsubsection{Domain Adaptation on Brain Tumor Segmentation}~\\
\textit{\textbf{UPenn $\rightarrow$ Others:}}. We further evaluate our method in a multi-class tumor segmentation task. We take 92 subjects from centre \emph{Upenn} and perform UDA on 50 scans from other centres. From Tab. \ref{tab:results_3}, we observe improvement on all the three classes. The improvement is not as large as in task 1 and 2. This could be attributed to the structure of a brain tumour which is much larger and complex than WMH and MS lesions. Although the improvement is limited, our method holds its promise of improving well-established state-of-the-art architectures.

\begin{table}[ht]
  \caption[table: Results Tumor]{Results on cross-centre brain tumor segmentation task. The values are calculated by averaging the results of the target-domain subjects. We compare our method with the performance before UDA (denoted by \emph{baseline}). TC=tumor core; e=edema; ET=enhancing tumor. Statistics analysis is done on the average of Dice scores of three tissues.}\label{tab:results_3}
  \centering
   \begin{tabular}{c c c c c}
    \hline
      Method &Dice$_{TC}$&Dice$_{e}$ & Dice$_{ET}$ & p-value$_{Dice}$ \\
    \hline
      Baseline &0.610 & 0.785& 0.796 & 0.026\\
     Ours (with a few shots) &~{0.622} & 0.792& 0.806 & 0.258\\
      Ours (with full set) &~\textbf{0.628} & \textbf{0.799}& \textbf{0.812} & -\\
    \hline
  \end{tabular}
\end{table}

\vspace{-0.2cm}

\subsubsection{Performance on both domains} \label{two_domains}
\begin{figure}[t]
	\begin{center}
		\includegraphics[width=0.48\textwidth,height=0.32\textwidth]{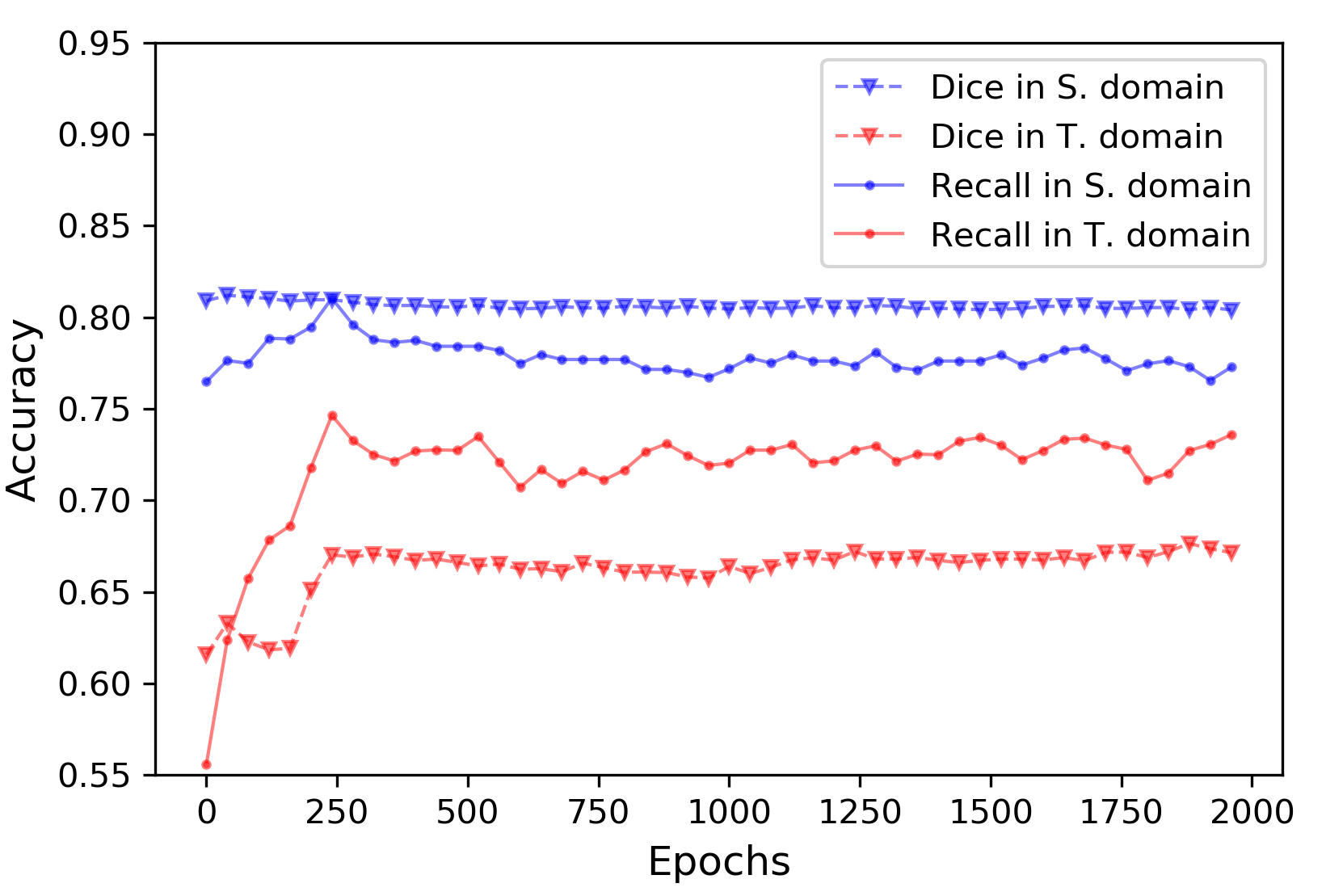}
	\end{center}
    	\caption{Performance on both domains during the domain adaptation process. We observe that the performance in the source domain remains stable whilst the performance in the target domain is increasing rapidly in the first 250 epochs. When the number of epochs reached 255, the variance and difference was smaller than 0.1 and larger the 6.1 respectively, and the UDA process is suggested to be stopped.}
    \label{fig:two_domains}
    \vspace{-0.1cm}
\end{figure}

To enforce the segmentation network not to forget the knowledge from the source domain, we train the segmentation network in a continual-learning manner using labeled data during the adaptation process. 
However, without a regularization, the segmentation model faces the risk of overfitting on the source domain since it was optimised in the first stage (i.e. pre-training) in Algorithm \ref{euclid}. We claim that the semantic discriminator regularizes the training in two aspects: 1) discriminating if a segmentation is good or not; 2) avoiding overfitting through the adversarial training.
We further perform a study to observe the behaviour of the segmentation model on both source domain and target domain without stopping the adaptation process.
In this experiment, we split the source-domain data of WMH segmentation task (Utrecht + Amsterdam $\rightarrow$ Singapore) into a training set (80\%) and a validation set (20\%) for observation. We use Dice score and lesion-wise recall for evaluation metrics. From Fig. \ref{fig:two_domains}, we confirm that the segmentation model does not overfit on the source domain whilst the performance on target domain is increasing stable in the first 250 epochs.

\subsubsection{Effect of the two new metrics} \label{metrics}

\vspace{-0.1cm}
\begin{figure}[t]
	\begin{center}
		\includegraphics[width=0.48\textwidth,height=0.32\textwidth]{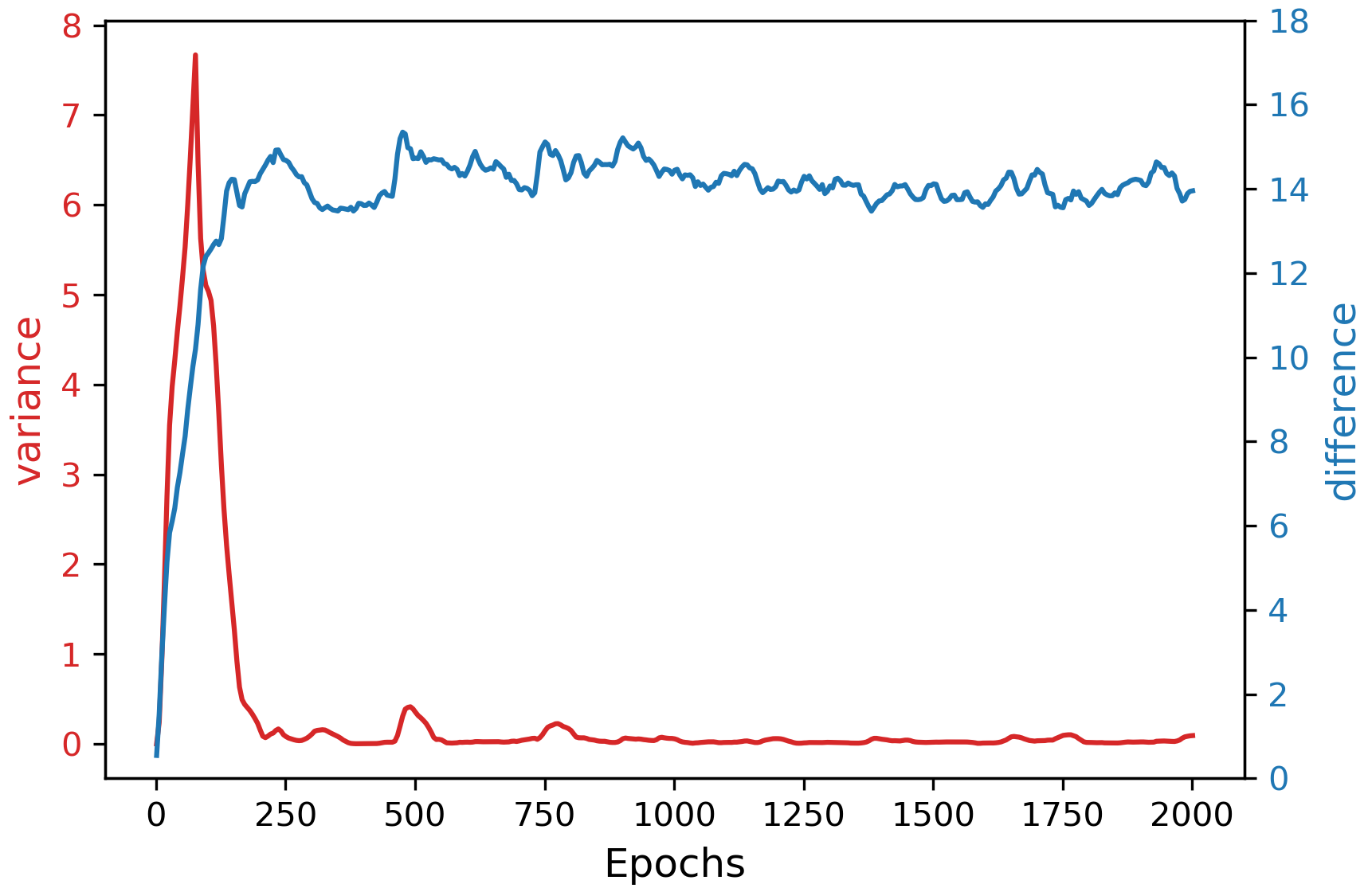}
	\end{center}
	\vspace{-0.2cm}
    	\caption{Plots of the two metrics over training epochs corresponding to Fig. \ref{fig:two_domains}. We found that the metrics are highly effective as a stopping criteria of UDA process without a validation set. The stopping point (epoch 255) corresponds to almost global maximum performance on the target domain.}
    \label{fig:var_diff}
    \vspace{-0.2cm}
\end{figure}


We perform an analysis to demonstrate the effectiveness of the proposed two metrics: variance and difference of predictions, without using a validation set from the target domain. In Fig. \ref{fig:var_diff}, We plot the curves of the two metrics during the UDA process in the above sub-section \ref{two_domains}. Based on the curves of two metrics, in the above experiment we stop the training when the variance is smaller than 0.1 and difference is larger than 6 which results in 255 epochs. We observe that it corresponds to almost global maximum performance on the target domain in Fig. \ref{fig:two_domains}. 

\begin{figure*}[ht]
	\begin{center}
		\includegraphics[width=0.76\textwidth,height=0.38\textwidth]{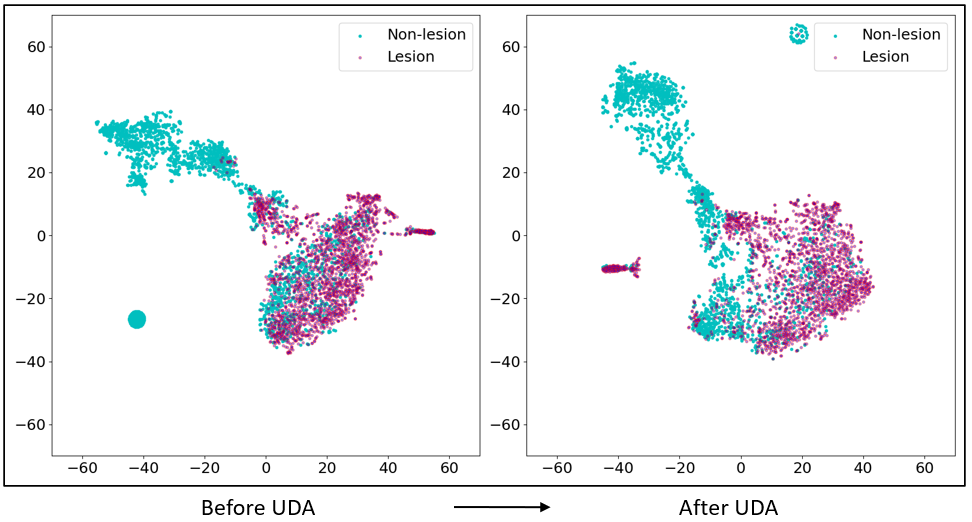}
	\end{center}
    	\caption{Illustration of pixel-wise feature distribution before and after UDA. (a) The distributions of lesion and non-lesion pixel-wise features. We observe that the lesion and non-lesion features are more separable after UDA. (b) Visualisation of lesion features in target domain before and after UDA. We found that the UDA process greatly transforms the target domain feature which are separable with the original ones. Visualization is done by t-SNE \cite{maaten2008visualizing}. (Best viewed in color)}
    \label{fig:t_SNE}
\end{figure*}

\vspace{-0.1cm}
\subsection{Ablation Study on Semantic and Boundary-aware Layers} \label{ablation}
We conduct an ablation experiment on WMH segmentation task (Utrecht + Amsterdam $\rightarrow$ Singapore) to evaluate the effectiveness of each key component in our proposed framework. Table. \ref{tab:ablation} shows the segmentation performance is increasingly better when more spatial information being included, especially, when incorporating edge and inverse maps. Similar trends were observed on other settings.

\begin{table}[ht]
  \caption[table: ablation]{Ablation study on key components including two semantic and boundary-aware layers.}\label{tab:ablation}
  \centering
  \begin{tabular}{c c c c c c}
    \toprule
      Methods & Mask & Edge Map & Inverse Map & Dice & H95$\downarrow$\\
    \midrule
    W$\setminus$o UDA &  & & &~67.4\% & 10.55\\
       Mask  & $\checkmark$  & & &~68.1\% &  9.72\\
      Edge Map & $\checkmark$& $\checkmark$ &   &~71.1\% & 8.20\\
      Inverse Map    & $\checkmark$& $\checkmark$&  $\checkmark$ &~73.9\% &7.41\\
    \bottomrule
    \vspace{-0.3cm}
  \end{tabular}
\end{table}

\vspace{-0.2cm}
\subsection{Feature Visualisation}

In Fig. \ref{fig:t_SNE}, we visualise the 64-dimensional features before soft-max layer corresponding to the region of interests (i.e. lesions) from the source and target domains to interpret the UDA process in the WMH segmentation task (Utrecht + Amsterdam $\rightarrow$ Singapore). We randomly sample 2000 features from each stage for \emph{t-SNE} visualisation. In Fig. \ref{fig:t_SNE}, we see that the lesion and non-lesion features are more separable after UDA, demonstrating that lesion and background are more distinguishable. 

\vspace{-0.2cm}
\section{Discussion and Conclusion}
In summary, we presented an efficient unsupervised domain adaptation framework that enforces arbitrary segmentation networks to adapt better to new domains. The proposed model learns underlying disease-specific spatial patterns in an adversarial manner. We found that using a few shots of unlabeled images from the target domain can significantly improve the segmentation results. Two effective metrics to measure the difference and variance are introduced to monitor the training process and stop it at a reasonable point.

The first finding is that the well-established segmentation networks (both 2D and 3D) can be further improved on new target domains by an adversarial optimization. The learning of spatial pattern distributions benefits from the semantic and boundary-aware layers. From Fig. \ref{fig:two_domains}, we observed that the lesion recall increases rapidly in the first 250 epochs indicating that the spatial patterns are captured by the segmentation network. 
The second finding is that the performances of the segmentation network on the source domains are maintained and the model does not suffer from over-fitting issues. Lastly, through the feature visualization, we see that the lesion and non-lesion features are more separable after UDA.

{
\bibliographystyle{IEEEtran}
\bibliography{egbib}

\begin{thebibliography}{10}
\providecommand{\url}[1]{#1}
\csname url@samestyle\endcsname
\providecommand{\newblock}{\relax}
\providecommand{\bibinfo}[2]{#2}
\providecommand{\BIBentrySTDinterwordspacing}{\spaceskip=0pt\relax}
\providecommand{\BIBentryALTinterwordstretchfactor}{4}
\providecommand{\BIBentryALTinterwordspacing}{\spaceskip=\fontdimen2\font plus
\BIBentryALTinterwordstretchfactor\fontdimen3\font minus
  \fontdimen4\font\relax}
\providecommand{\BIBforeignlanguage}[2]{{%
\expandafter\ifx\csname l@#1\endcsname\relax
\typeout{** WARNING: IEEEtran.bst: No hyphenation pattern has been}%
\typeout{** loaded for the language `#1'. Using the pattern for}%
\typeout{** the default language instead.}%
\else
\language=\csname l@#1\endcsname
\fi
#2}}
\providecommand{\BIBdecl}{\relax}
\BIBdecl

\bibitem{litjens2017survey}
G.~Litjens, T.~Kooi, B.~E. Bejnordi, A.~A.~A. Setio, F.~Ciompi, M.~Ghafoorian,
  J.~A. Van Der~Laak, B.~Van~Ginneken, and C.~I. S{\'a}nchez, ``A survey on
  deep learning in medical image analysis,'' \emph{Medical image analysis},
  vol.~42, pp. 60--88, 2017.

\bibitem{ronneberger2015u}
O.~Ronneberger, P.~Fischer, and T.~Brox, ``U-net: Convolutional networks for
  biomedical image segmentation,'' in \emph{International Conference on Medical
  image computing and computer-assisted intervention}.\hskip 1em plus 0.5em
  minus 0.4em\relax Springer, 2015, pp. 234--241.

\bibitem{kamnitsas2017efficient}
K.~Kamnitsas, C.~Ledig, V.~F. Newcombe, J.~P. Simpson, A.~D. Kane, D.~K. Menon,
  D.~Rueckert, and B.~Glocker, ``Efficient multi-scale 3d cnn with fully
  connected crf for accurate brain lesion segmentation,'' \emph{Medical Image
  Analysis}, vol. 100, no.~36, pp. 61--78, 2017.

\bibitem{ben2010theory}
S.~Ben-David, J.~Blitzer, K.~Crammer, A.~Kulesza, F.~Pereira, and J.~W.
  Vaughan, ``A theory of learning from different domains,'' \emph{Machine
  learning}, vol.~79, no. 1-2, pp. 151--175, 2010.

\bibitem{glocker2019machine}
B.~Glocker, R.~Robinson, D.~C. Castro, Q.~Dou, and E.~Konukoglu, ``Machine
  learning with multi-site imaging data: An empirical study on the impact of
  scanner effects,'' \emph{arXiv preprint arXiv:1910.04597}, 2019.

\bibitem{karani2018lifelong}
N.~Karani, K.~Chaitanya, C.~Baumgartner, and E.~Konukoglu, ``A lifelong
  learning approach to brain mr segmentation across scanners and protocols,''
  in \emph{International Conference on Medical Image Computing and
  Computer-Assisted Intervention}.\hskip 1em plus 0.5em minus 0.4em\relax
  Springer, 2018, pp. 476--484.

\bibitem{pan2009survey}
S.~J. Pan and Q.~Yang, ``A survey on transfer learning,'' \emph{IEEE
  Transactions on knowledge and data engineering}, vol.~22, no.~10, pp.
  1345--1359, 2009.

\bibitem{ganin2014unsupervised}
Y.~Ganin and V.~Lempitsky, ``Unsupervised domain adaptation by
  backpropagation,'' \emph{arXiv preprint arXiv:1409.7495}, 2014.

\bibitem{jager2008nonrigid}
F.~Jager and J.~Hornegger, ``Nonrigid registration of joint histograms for
  intensity standardization in magnetic resonance imaging,'' \emph{IEEE
  Transactions on Medical Imaging}, vol.~28, no.~1, pp. 137--150, 2008.

\bibitem{long2015learning}
M.~Long, Y.~Cao, J.~Wang, and M.~I. Jordan, ``Learning transferable features
  with deep adaptation networks,'' in \emph{Proceedings of the 32nd
  International Conference on International Conference on Machine
  Learning-Volume 37}.\hskip 1em plus 0.5em minus 0.4em\relax JMLR. org, 2015,
  pp. 97--105.

\bibitem{long2016unsupervised}
M.~Long, H.~Zhu, J.~Wang, and M.~I. Jordan, ``Unsupervised domain adaptation
  with residual transfer networks,'' in \emph{Advances in Neural Information
  Processing Systems}, 2016, pp. 136--144.

\bibitem{goodfellow2014generative}
I.~Goodfellow, J.~Pouget-Abadie, M.~Mirza, B.~Xu, D.~Warde-Farley, S.~Ozair,
  A.~Courville, and Y.~Bengio, ``Generative adversarial nets,'' in
  \emph{Advances in neural information processing systems}, 2014, pp.
  2672--2680.

\bibitem{arjovsky2017wasserstein}
M.~Arjovsky, S.~Chintala, and L.~Bottou, ``Wasserstein gan,'' \emph{arXiv
  preprint arXiv:1701.07875}, 2017.

\bibitem{ganin2016domain}
Y.~Ganin, E.~Ustinova, H.~Ajakan, P.~Germain, H.~Larochelle, F.~Laviolette,
  M.~Marchand, and V.~Lempitsky, ``Domain-adversarial training of neural
  networks,'' \emph{The Journal of Machine Learning Research}, vol.~17, no.~1,
  pp. 2096--2030, 2016.

\bibitem{tzeng2017adversarial}
E.~Tzeng, J.~Hoffman, K.~Saenko, and T.~Darrell, ``Adversarial discriminative
  domain adaptation,'' in \emph{Proceedings of the IEEE Conference on Computer
  Vision and Pattern Recognition}, 2017, pp. 7167--7176.

\bibitem{zhang2017curriculum}
Y.~Zhang, P.~David, and B.~Gong, ``Curriculum domain adaptation for semantic
  segmentation of urban scenes,'' in \emph{Proceedings of the IEEE
  International Conference on Computer Vision}, 2017, pp. 2020--2030.

\bibitem{hoffman2016fcns}
J.~Hoffman, D.~Wang, F.~Yu, and T.~Darrell, ``Fcns in the wild: Pixel-level
  adversarial and constraint-based adaptation,'' \emph{arXiv preprint
  arXiv:1612.02649}, 2016.

\bibitem{ouyang2019data}
C.~Ouyang, K.~Kamnitsas, C.~Biffi, J.~Duan, and D.~Rueckert, ``Data efficient
  unsupervised domain adaptation for cross-modality image segmentation,'' in
  \emph{International Conference on Medical Image Computing and
  Computer-Assisted Intervention}.\hskip 1em plus 0.5em minus 0.4em\relax
  Springer, 2019, pp. 669--677.

\bibitem{orbes2019knowledge}
M.~Orbes-Arteainst, J.~Cardoso, L.~S{\o}rensen, C.~Igel, S.~Ourselin, M.~Modat,
  M.~Nielsen, and A.~Pai, ``Knowledge distillation for semi-supervised domain
  adaptation,'' in \emph{OR 2.0 Context-Aware Operating Theaters and Machine
  Learning in Clinical Neuroimaging}.\hskip 1em plus 0.5em minus 0.4em\relax
  Springer, 2019, pp. 68--76.

\bibitem{kamnitsas2017unsupervised}
K.~Kamnitsas, C.~Baumgartner, C.~Ledig, V.~Newcombe, J.~Simpson, A.~Kane,
  D.~Menon, A.~Nori, A.~Criminisi, D.~Rueckert \emph{et~al.}, ``Unsupervised
  domain adaptation in brain lesion segmentation with adversarial networks,''
  in \emph{International conference on information processing in medical
  imaging}.\hskip 1em plus 0.5em minus 0.4em\relax Springer, 2017, pp.
  597--609.

\bibitem{dou2018unsupervised}
Q.~Dou, C.~Ouyang, C.~Chen, H.~Chen, and P.-A. Heng, ``Unsupervised
  cross-modality domain adaptation of convnets for biomedical image
  segmentations with adversarial loss,'' in \emph{Proceedings of the 27th
  International Joint Conference on Artificial Intelligence}, 2018, pp.
  691--697.

\bibitem{zhao2018supervised}
H.~Zhao, H.~Li, S.~Maurer-Stroh, Y.~Guo, Q.~Deng, and L.~Cheng, ``Supervised
  segmentation of un-annotated retinal fundus images by synthesis,'' \emph{IEEE
  transactions on medical imaging}, vol.~38, no.~1, pp. 46--56, 2018.

\bibitem{huo2018adversarial}
Y.~Huo, Z.~Xu, S.~Bao, A.~Assad, R.~G. Abramson, and B.~A. Landman,
  ``Adversarial synthesis learning enables segmentation without target modality
  ground truth,'' in \emph{2018 IEEE 15th International Symposium on Biomedical
  Imaging (ISBI 2018)}.\hskip 1em plus 0.5em minus 0.4em\relax IEEE, 2018, pp.
  1217--1220.

\bibitem{bousmalis2017unsupervised}
K.~Bousmalis, N.~Silberman, D.~Dohan, D.~Erhan, and D.~Krishnan, ``Unsupervised
  pixel-level domain adaptation with generative adversarial networks,'' in
  \emph{Proceedings of the IEEE conference on computer vision and pattern
  recognition}, 2017, pp. 3722--3731.

\bibitem{chen2019synergistic}
C.~Chen, Q.~Dou, H.~Chen, J.~Qin, and P.-A. Heng, ``Synergistic image and
  feature adaptation: Towards cross-modality domain adaptation for medical
  image segmentation,'' \emph{arXiv preprint arXiv:1901.08211}, 2019.

\bibitem{zhang2020generalizing}
L.~Zhang, X.~Wang, D.~Yang, T.~Sanford, S.~Harmon, B.~Turkbey, B.~J. Wood,
  H.~Roth, A.~Myronenko, D.~Xu \emph{et~al.}, ``Generalizing deep learning for
  medical image segmentation to unseen domains via deep stacked
  transformation,'' \emph{IEEE Transactions on Medical Imaging}, 2020.

\bibitem{goodfellow2013empirical}
I.~J. Goodfellow, M.~Mirza, D.~Xiao, A.~Courville, and Y.~Bengio, ``An
  empirical investigation of catastrophic forgetting in gradient-based neural
  networks,'' \emph{arXiv preprint arXiv:1312.6211}, 2013.

\bibitem{ozdemir2018learn}
F.~Ozdemir, P.~Fuernstahl, and O.~Goksel, ``Learn the new, keep the old:
  Extending pretrained models with new anatomy and images,'' in
  \emph{International Conference on Medical Image Computing and
  Computer-Assisted Intervention}.\hskip 1em plus 0.5em minus 0.4em\relax
  Springer, 2018, pp. 361--369.

\bibitem{ding2019boundary}
H.~Ding, X.~Jiang, A.~Q. Liu, N.~M. Thalmann, and G.~Wang, ``Boundary-aware
  feature propagation for scene segmentation,'' in \emph{Proceedings of the
  IEEE International Conference on Computer Vision}, 2019, pp. 6819--6829.

\bibitem{shen2017boundary}
H.~Shen, R.~Wang, J.~Zhang, and S.~J. McKenna, ``Boundary-aware fully
  convolutional network for brain tumor segmentation,'' in \emph{International
  Conference on Medical Image Computing and Computer-Assisted
  Intervention}.\hskip 1em plus 0.5em minus 0.4em\relax Springer, 2017, pp.
  433--441.

\bibitem{gao2010improved}
W.~Gao, X.~Zhang, L.~Yang, and H.~Liu, ``An improved sobel edge detection,'' in
  \emph{2010 3rd International Conference on Computer Science and Information
  Technology}, vol.~5.\hskip 1em plus 0.5em minus 0.4em\relax IEEE, 2010, pp.
  67--71.

\bibitem{li2018fully}
H.~Li, G.~Jiang, J.~Zhang, R.~Wang, Z.~Wang, W.-S. Zheng, and B.~Menze, ``Fully
  convolutional network ensembles for white matter hyperintensities
  segmentation in mr images,'' \emph{NeuroImage}, vol. 183, pp. 650--665, 2018.

\bibitem{wang2017automatic}
G.~Wang, W.~Li, S.~Ourselin, and T.~Vercauteren, ``Automatic brain tumor
  segmentation using cascaded anisotropic convolutional neural networks,'' in
  \emph{International MICCAI brainlesion workshop}.\hskip 1em plus 0.5em minus
  0.4em\relax Springer, 2017, pp. 178--190.

\bibitem{milletari2016v}
F.~Milletari, N.~Navab, and S.-A. Ahmadi, ``V-net: Fully convolutional neural
  networks for volumetric medical image segmentation,'' in \emph{2016 Fourth
  International Conference on 3D Vision (3DV)}.\hskip 1em plus 0.5em minus
  0.4em\relax IEEE, 2016, pp. 565--571.

\bibitem{bakas2018identifying}
S.~Bakas, M.~Reyes, A.~Jakab, S.~Bauer, M.~Rempfler, A.~Crimi, R.~T. Shinohara,
  C.~Berger, S.~M. Ha, M.~Rozycki \emph{et~al.}, ``Identifying the best machine
  learning algorithms for brain tumor segmentation, progression assessment, and
  overall survival prediction in the brats challenge,'' \emph{arXiv preprint
  arXiv:1811.02629}, 2018.

\bibitem{kuijf2019standardized}
H.~J. Kuijf, J.~M. Biesbroek, J.~de~Bresser, R.~Heinen, S.~Andermatt, M.~Bento,
  M.~Berseth, M.~Belyaev, M.~J. Cardoso, A.~Casamitjana \emph{et~al.},
  ``Standardized assessment of automatic segmentation of white matter
  hyperintensities; results of the wmh segmentation challenge,'' \emph{IEEE
  transactions on medical imaging}, 2019.

\bibitem{zhuang2019evaluation}
X.~Zhuang, L.~Li, C.~Payer, D.~Stern, M.~Urschler, M.~P. Heinrich, J.~Oster,
  C.~Wang, O.~Smedby, C.~Bian \emph{et~al.}, ``Evaluation of algorithms for
  multi-modality whole heart segmentation: An open-access grand challenge,''
  \emph{arXiv preprint arXiv:1902.07880}, 2019.

\bibitem{carass2017longitudinal}
A.~Carass, S.~Roy, A.~Jog, J.~L. Cuzzocreo, E.~Magrath, A.~Gherman, J.~Button,
  J.~Nguyen, F.~Prados, C.~H. Sudre \emph{et~al.}, ``Longitudinal multiple
  sclerosis lesion segmentation: resource and challenge,'' \emph{NeuroImage},
  vol. 148, pp. 77--102, 2017.

\bibitem{menze2014multimodal}
B.~H. Menze, A.~Jakab, S.~Bauer, J.~Kalpathy-Cramer, K.~Farahani, J.~Kirby,
  Y.~Burren, N.~Porz, J.~Slotboom, R.~Wiest \emph{et~al.}, ``The multimodal
  brain tumor image segmentation benchmark (brats),'' \emph{IEEE transactions
  on medical imaging}, vol.~34, no.~10, pp. 1993--2024, 2014.

\bibitem{hoffman2017cycada}
J.~Hoffman, E.~Tzeng, T.~Park, J.-Y. Zhu, P.~Isola, K.~Saenko, A.~A. Efros, and
  T.~Darrell, ``Cycada: Cycle-consistent adversarial domain adaptation,''
  \emph{arXiv preprint arXiv:1711.03213}, 2017.

\bibitem{isola2017image}
P.~Isola, J.-Y. Zhu, T.~Zhou, and A.~A. Efros, ``Image-to-image translation
  with conditional adversarial networks,'' in \emph{Proceedings of the IEEE
  conference on computer vision and pattern recognition}, 2017, pp. 1125--1134.

\bibitem{kingma2014adam}
D.~P. Kingma and J.~Ba, ``Adam: A method for stochastic optimization,''
  \emph{arXiv preprint arXiv:1412.6980}, 2014.

\bibitem{zhu2017unpaired}
J.-Y. Zhu, T.~Park, P.~Isola, and A.~A. Efros, ``Unpaired image-to-image
  translation using cycle-consistent adversarial networks,'' in
  \emph{Proceedings of the IEEE international conference on computer vision},
  2017, pp. 2223--2232.

\bibitem{maaten2008visualizing}
L.~v.~d. Maaten and G.~Hinton, ``Visualizing data using t-sne,'' \emph{Journal
  of machine learning research}, vol.~9, no. Nov, pp. 2579--2605, 2008.

\end{thebibliography}
}

\end{document}